\documentclass[11pt,a4paper]{article}
\usepackage[margin=25mm]{geometry}
\usepackage[T1]{fontenc}
\usepackage[utf8]{inputenc}
\usepackage{lmodern,microtype,graphicx,booktabs,amsmath,amssymb}
\usepackage{float}
\usepackage[font=small,labelfont=bf]{caption}
\usepackage[hidelinks]{hyperref}
\usepackage[authoryear,round]{natbib}
\usepackage{enumitem}
\title{Reinforcement Learning for Adaptive Crowd Flow Control\\Using Dynamic Barriers}

\author{
Alhanof Alolyan$^{1,*}$,
He Wang$^{2}$,
Amani Alonayzan$^{3}$\\[1em]
\small $^{1}$Princess Nourah bint Abdulrahman University, Saudi Arabia\\
\small $^{2}$University College London, United Kingdom\\
\small $^{3}$Independent Researcher\\[1em]
\small $^{1}$\texttt{hano.alolyan@gmail.com},
\texttt{ASAlolyan@pnu.edu.sa}\\
\small $^{2}$\texttt{he\_wang@ucl.ac.uk}\\
\small $^{3}$\texttt{Amani.onayzan@gmail.com}\\[1em]
\small $^{*}$Corresponding author: Alhanof Alolyan
}

\begin{document}
\maketitle
\begin{quote}\small\textbf{Publication note.} An earlier version of this work was presented at the 26th Scientific Forum for Hajj, Umrah and Visit Research, organised by Umm Al-Qura University, Makkah, in 2026. This author manuscript includes minor revisions and is shared to make the research easier to access and cite.\end{quote}
\begin{abstract}
Crowd barriers are widely used to regulate pedestrian movement during large public events such as pilgrimages. However, fixed barrier layouts often create unnecessary stopping and local congestion even when they successfully control the overall crowd flow. Finding better barrier configurations is difficult because there is no dataset that provides the best barrier placement for a given crowd scene. We address this challenge by introducing a simulation-driven reinforcement learning framework that learns barrier configurations without requiring labelled training data. Starting from a segmented crowd scene, we construct a realistic simulation environment that reproduces the observed pedestrian distribution. This environment is then used to train a Soft Actor-Critic (SAC) agent that continuously adjusts barrier positions to improve pedestrian movement. Unlike supervised approaches, the proposed framework allows the agent to explore barrier configurations that have never been observed in real data. We evaluate the learned policy under unseen numbers of pedestrians. The learned policy consistently outperforms the existing manually designed barrier configuration by reducing pedestrian stops while maintaining effective flow regulation. These results demonstrate that combining computer vision, crowd simulation, and reinforcement learning provides an effective framework for optimizing crowd control infrastructure. The source code is available on GitHub: \url{https://github.com/HanoCat/RLForHajjProject}
\end{abstract}
\section{Introduction}
Crowd barriers are commonly used to regulate pedestrian movement in dense public spaces. However, manually designed layouts can create unnecessary stopping behaviour and local congestion. This is especially problematic when barriers are intended to regulate the overall crowd flow while avoiding inefficient movement locally. Existing crowd simulation methods mainly focus on modelling pedestrian behaviour or providing interactive tools for manually designing crowd control layouts rather than learning barrier configurations automatically \citep{gonzalez2021,charalambous2023}. Reinforcement learning has also been applied to crowd simulation and evacuation planning by learning pedestrian behaviours or tuning simulation parameters \citep{zhao2022,zhang2024}. However, learning how to optimize physical barrier configurations from an observed crowd scene has less attention.

We propose a simulation-driven framework for adaptive crowd barrier control. Starting from a segmented crowd image, we construct a pedestrian simulation that reproduces the observed crowd distribution. This simulation serves as the environment for Soft Actor-Critic (SAC) that learns continuous barrier configurations. The reward encourages smooth pedestrian movement by reducing stopping behaviour and local congestion while preserving the intended flow regulation of the barriers. Unlike supervised learning, our framework does not require labelled examples of optimal barrier placements and allows exploration beyond manually designed layouts. Experiments demonstrate that the learned policy consistently outperforms the existing ground-truth configuration and generalizes to unseen crowd sizes.
\section{Methodology}
\begin{figure}[htbp]\centering\includegraphics[width=\textwidth]{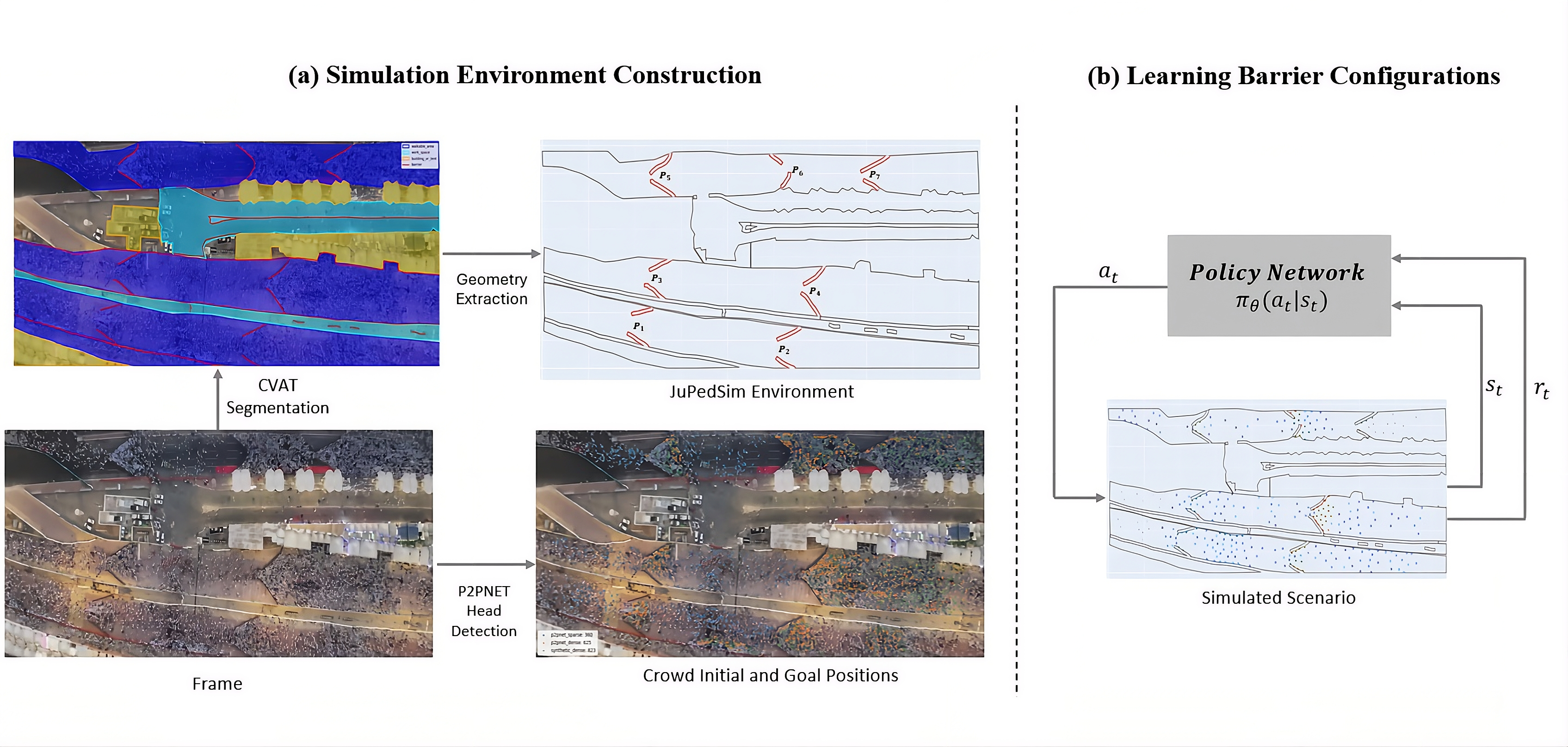}
\caption{Overview of the approach. (a) CVAT annotations and P2PNet pedestrian detections are used to construct a JuPedSim simulation. (b) A reinforcement learning policy adjusts barriers and receives feedback from the simulated crowd.}\label{fig:framework}\end{figure}

Figure~\ref{fig:framework} summarises the two stages of the approach.
Figure 1 illustrates the proposed framework, which consists of two main stages. First, we annotate a crowd image to identify the walkable regions, obstacles, and barrier locations using CVAT \citep{sekachev2019}. The extracted geometry is converted into a JuPedSim simulation environment \citep{wagoum2015}. Pedestrian head locations are detected using P2PNet \citep{song2021} and each pedestrian is assigned an initial position and a destination according to the segmented scene. This process produces a realistic simulation that preserves the observed crowd distribution while allowing different barrier configurations to be evaluated. Unlike collecting real barrier-control data, the simulation provides unlimited interactions for reinforcement learning.

Second, we train a Soft Actor-Critic (SAC) agent in the constructed simulation. At time step $t$, the state contains the number of pedestrians $N$, the barrier configuration $B_t$, and the pedestrian state $P_t$:
\begin{equation}\label{eq:state}
s_t=\{N,B_t,P_t\},\qquad P_t=\{(x_i,y_i,v_i)\}_{i=1}^{N}.
\end{equation}
Here, $(x_i,y_i)$ is a pedestrian's position and $v_i$ is their speed. The action $a_t$ adjusts the translation $(\Delta x,\Delta y)$ and rotation $\Delta\theta$ of each barrier pair. The reward penalises stopping and slow movement:
\begin{equation}\label{eq:reward}
r_t=-C,\qquad C=\lambda_s R_{\mathrm{stop}}+\lambda_v\left(1-\frac{\bar v}{v_{\max}}\right),\qquad \lambda_s+\lambda_v=1.
\end{equation}
In this expression, $R_{\mathrm{stop}}$ is the proportion of stopped pedestrians, $\bar v$ is the mean pedestrian speed, and $v_{\max}$ is the desired maximum walking speed. Lower stopping and higher walking speeds lead to a better reward.
\section{Results and Discussion}
\begin{figure}[htbp]\centering\includegraphics[width=.84\textwidth]{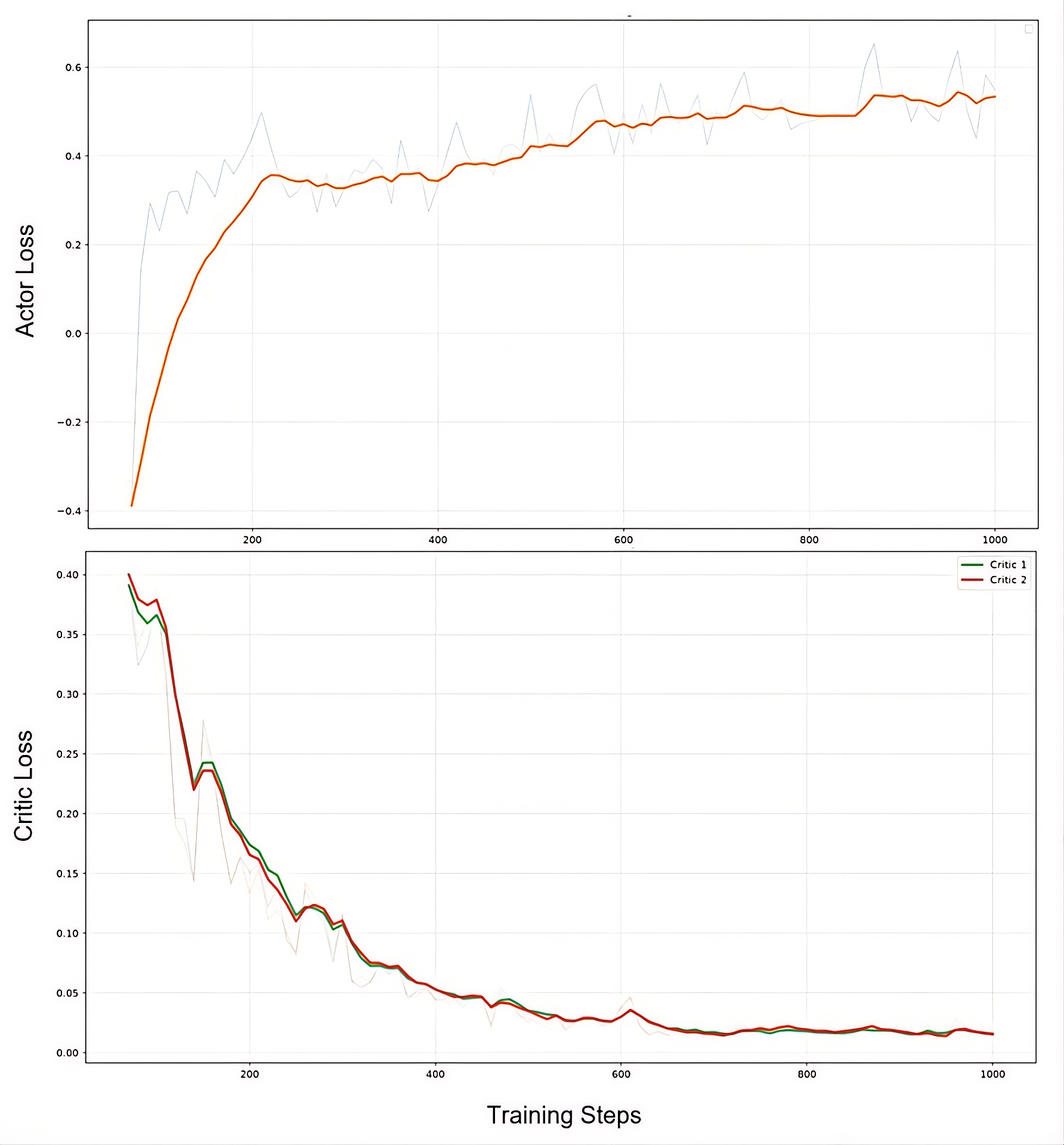}
\caption{SAC training curves. The top plot shows actor loss; the bottom plot shows losses of the two critics. Faint lines show raw values and solid lines show smoothed trends.}\label{fig:training}\end{figure}

Figure~\ref{fig:training} shows the training curves. The critic losses decrease during training, while the actor loss gradually stabilises. These curves describe optimisation behaviour; the comparison against other barrier configurations provides the main evidence of performance.
We compare the learned policy with the original ground-truth (GT) barriers, all-open barriers, and all-closed barriers. The all-open configuration allows pedestrians to move with fewer restrictions, giving the highest throughput and lowest stopping ratio overall. However, removing every barrier would not meet the intended purpose of regulating crowd movement. The all-closed configuration creates substantial congestion.
\begin{figure}[htbp]\centering\includegraphics[width=\textwidth]{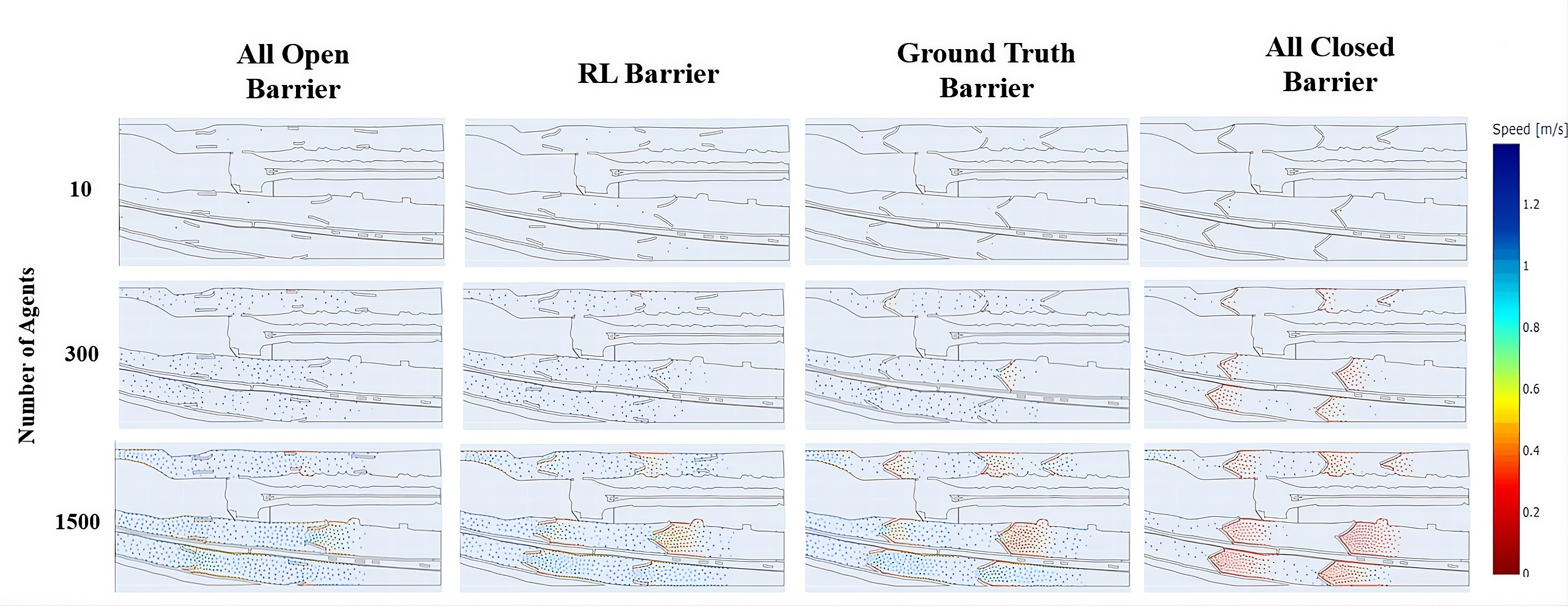}
\caption{Qualitative comparison of barrier configurations for 10, 300, and 1,500 pedestrians. The policy was trained on crowd sizes from 100 to 700; the 10- and 1,500-pedestrian cases are outside this training range. Colour indicates pedestrian speed.}\label{fig:barriers}\end{figure}
The learned policy provides a compromise between the original layout and the two extreme settings. As Table~\ref{tab:results} shows, it improves throughput and mean normalised speed while reducing the stopped-pedestrian ratio relative to GT. The all-open baseline still achieves better flow metrics, so the learned policy should not be described as the best-performing configuration on every measure.
\begin{table}[htbp]\centering\small
\caption{Performance across evaluation scenarios (mean $\pm$ standard deviation). Higher throughput and speed are preferable; a lower stopped ratio is preferable.}\label{tab:results}
\begin{tabular}{lccc}\toprule
Configuration & Stopped ratio $\downarrow$ & Normalised speed $\uparrow$ & Throughput (agents/s) $\uparrow$\\\midrule
GT & $0.104\pm0.076$ & $0.887\pm0.223$ & $4.604\pm2.966$\\
All open & $0.035\pm0.023$ & $1.000\pm0.134$ & $5.447\pm3.544$\\
All closed & $0.642\pm0.075$ & $0.304\pm0.097$ & $1.991\pm1.287$\\
RL (ours) & $0.072\pm0.075$ & $0.942\pm0.155$ & $5.098\pm3.302$\\\bottomrule
\end{tabular}\end{table}
Figure~\ref{fig:metrics} provides the results for individual crowd sizes. These results include crowd sizes not seen during training. They suggest that the policy can transfer to different crowd sizes in the tested simulation, although broader testing would be needed to establish generalisation to other physical scenes.
\begin{figure}[htbp]\centering\includegraphics[width=\textwidth]{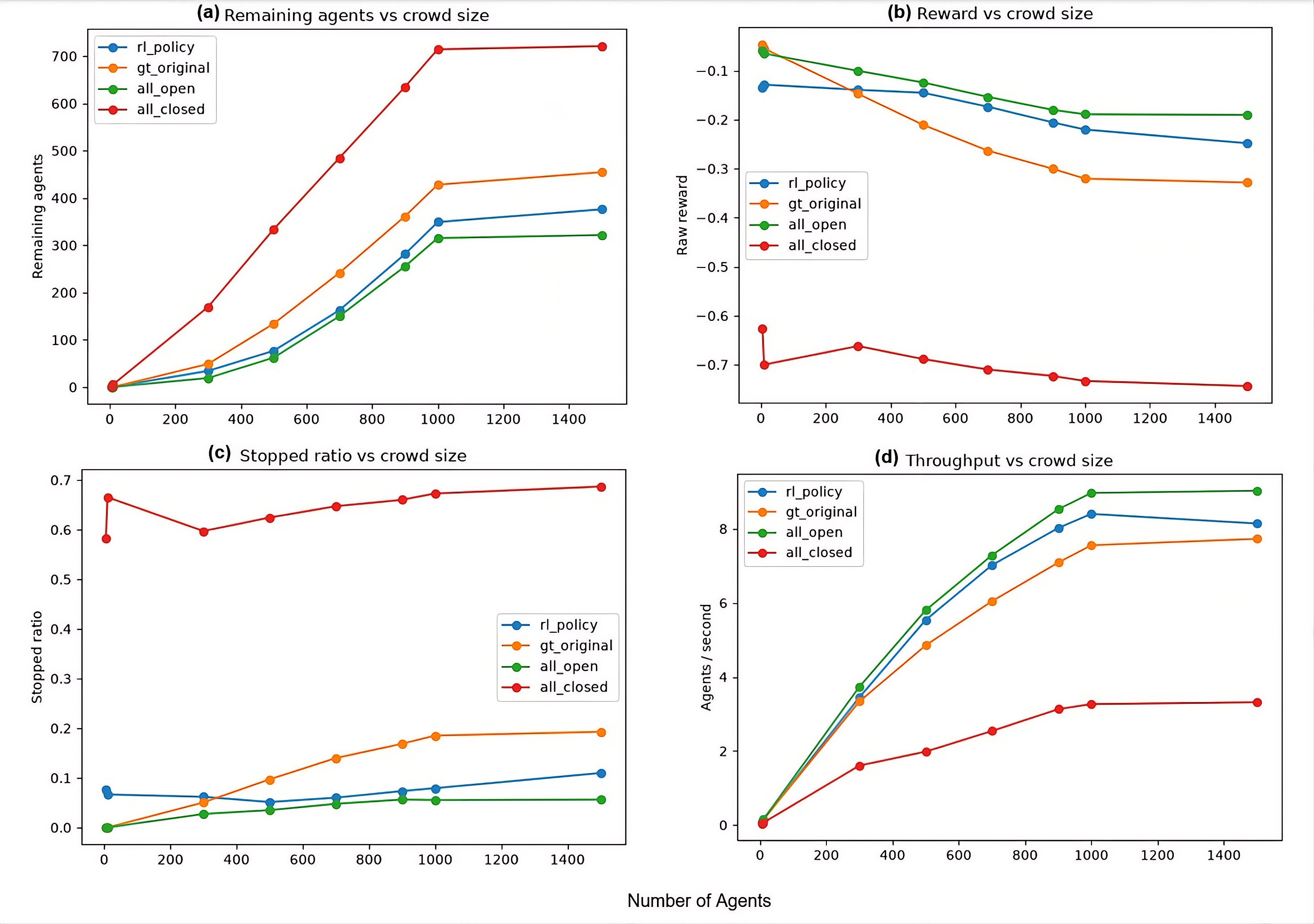}
\caption{Comparison by crowd size: (a) remaining pedestrians at the end of the simulation, (b) reward, (c) stopped ratio, and (d) throughput.}\label{fig:metrics}\end{figure}
\section{Conclusions}
We presented a simulation-driven method for learning barrier configurations without labelled examples of optimal layouts. The approach builds a crowd simulation from an observed scene and trains a SAC agent to adjust barriers based on simulated pedestrian movement. Relative to the original GT layout, the learned policy increased throughput from 4.604 to 5.098 agents/s (10.73\%), increased normalised speed from 0.887 to 0.942 (6.20\%), and reduced the stopped ratio from 0.104 to 0.072 (30.77\%). The results support the potential of reinforcement learning to improve pedestrian movement while retaining barriers for crowd regulation. These findings are based on the tested simulated environment and do not yet establish real-world performance.
\section{Future Work}
\begin{itemize}[leftmargin=*]
\item Improve scene calibration using known corridor widths, barrier dimensions, or site maps to convert image pixels into real-world distances.
\item Investigate faster simulation engines to reduce training cost.
\item Explore more sample-efficient reinforcement learning methods.
\item Test additional crowd scenes and live observations before considering real-world deployment.
\end{itemize}
\bibliographystyle{plainnat}
\bibliography{references}
\end{document}